 \documentclass[letterpaper,journal]{IEEEtran}

\ifCLASSINFOpdf
 
\else
 
\fi

\usepackage{graphics}
\usepackage{subfig}
\usepackage{epsfig} 
\usepackage{cite}
\usepackage{color}
\usepackage{amssymb}
\usepackage[font={small}]{caption}
\usepackage{amsmath}
\usepackage{algorithm}
\usepackage[noend]{algpseudocode}
\usepackage{multirow}

\usepackage{subcaption}

\usepackage{caption}
\usepackage{subcaption}
\usepackage{graphicx}
\usepackage{subfig}

\usepackage{forest}
\usepackage{tikz-qtree}
\usepackage{graphics}
\usepackage{subfig}
\usepackage{array}
\usepackage{longtable}
\usepackage{epsfig} 

\begin{document}
\title{ SHM-Traffic: DRL and Transfer learning based  UAV Control for Structural Health Monitoring of Bridges with Traffic } 

\author{Divija Swetha Gadiraju, Saeed Eftekhar Azam and Deepak Khazanchi
\thanks{D. S. Gadiraju,  and D. Khazanchi are with with University of Nebraska at Omaha. S.E. Azam is with University of New Hampshire.\ {This work is partially supported by contracts W912HZ21C0060 and W912HZ23C0005, U.S. Army Corps of Engineers, Engineering Research and Development Center (ERDC).}
\\ Email:  \{dgadiraju,khazanchi\}@unomaha.edu, saeed.eftekharazam@unh.edu \
}}

\newcommand{\ourmethod}{SHM-Traffic}


\maketitle

\begin{abstract}
This work focuses on using advanced techniques for structural health monitoring (SHM) for bridges with Traffic (\ourmethod). We propose an approach using deep reinforcement learning (DRL)-based control for Unmanned Aerial Vehicle (UAV). Our approach conducts concrete bridge deck survey while traffic is ongoing and detect cracks. The UAV performs the crack detection, and the location of cracks is initially unknown. We use two edge detection techniques. First, we use canny edge detection for crack detection. We also use a Convolutional Neural Network (CNN) for crack detection and compare it with canny edge detection. Transfer learning is applied using CNN with pre-trained weights obtained from a crack image dataset. This enables the model to adapt and improve its performance in identifying and localizing cracks. Proximal Policy Optimization (PPO) is applied for UAV control and bridge surveys.  The experimentation across various scenarios is performed to evaluate the performance of the proposed methodology. Key metrics such as task completion time and reward convergence are observed to gauge the effectiveness of the approach. We observe that the Canny edge detector offers up to 40\% lower task completion time, while the CNN excels in up to 12\% better damage detection and 1.8 times better rewards.

\end{abstract}



\section{Introduction}

Bridges play a critical role in modern infrastructure, facilitating transportation and connecting communities. However, the aging global bridge infrastructure poses challenges related to maintenance, safety, and longevity. Structural Health Monitoring (SHM) has emerged as a vital tool to assess and manage the condition of bridges, ensuring their reliability and safety over time \cite{mlshm20}. One innovative and efficient approach to conducting structural health assessments is through the integration of Unmanned Aerial Vehicles (UAVs) or drones. Unmanned Aerial Vehicles (UAVs) or drones have completely transformed the landscape of SHM \cite{mlshm21}. New sensors and imaging have replaced old inspection methods, providing more accurate and efficient results \cite{uavshm}. UAVs provide a rapid, cost-effective, and all-encompassing means of data collection. They capture high-resolution images, perform thermal imaging, and can access difficult-to-reach areas without posing any risks to personnel safety \cite{mlshm22}.
Deep reinforcement learning (DRL) is a machine learning technique that combines deep learning's perception ability with reinforcement learning's decision-making ability to train AI systems to make better decisions through experience \cite{mnih2015human}. This work employs a DRL-based UAV control system for bridge surveying and crack detection. The significance of SHM is addressed by using DRL-based UAVs in bridge inspections, which allows for real-time monitoring, early detection of structural defects, and timely decision-making to ensure the safety and longevity of these critical infrastructure assets \cite{zhao2021structural}. 

In the proposed approach, SHM-Traffic, bridge surveys are conducted while traffic is present on the bridge, enhancing the practicality and real-time applicability of the methodology. The UAV is equipped with a camera, and the location of cracks is initially unknown. To address this challenge, an edge detection technique is employed for crack detection \cite{edgedet}. Subsequently, another approach, incorporating transfer learning \cite{zhu2023transfer} is used for crack detection. Transfer learning is applied using a Convolutional Neural Network (CNN) with pre-trained weights obtained from a crack image dataset. This enables the model to adapt and improve its performance in identifying and localizing cracks. In \ourmethod, Proximal Policy Optimization (PPO) \cite{schulman2017proximal}, one of the famous DRL techniques, is leveraged for both UAV control and bridge survey tasks.  This research can be generalized to any concrete structure. We consider bridges for our analysis. Extensive experimentation is conducted across various scenarios to evaluate the performance of the proposed methodology. Key metrics such as task completion time and reward convergence are observed to gauge the effectiveness of the approach. The canny edge detector exhibits a lower task completion time compared to CNN. However, CNN demonstrates superior damage detection capabilities, achieving higher rewards. This trade-off between task completion time and damage detection accuracy provides valuable insights for optimizing the UAV-based SHM system in practical applications.

The main contributions of this work are as follows:
\begin{enumerate}
    \item The bridge survey using UAV control problem is formulated as a Markov Decision Process (MDP). The state space takes into account the local situational information. The location of the cracks or the damage level of the bridge is unknown to the UAV before the time of operation. 
    \item We propose two crack detection techniques to identify cracks from the UAV collected image data. After the crack detection, the exact crack locations are determined. We use a canny edge detector and a CNN. The CNN uses a pre-trained network which is trained on concrete crack image dataset using Transfer learning.
    \item We build a concrete bridge simulator for our experimentation. To the author's knowledge, this is the first work proposing the survey of bridges with traffic. The uninterrupted traffic is present on the bridge throughout the UAV survey and crack detection.
    \item We use a PPO-based approach called \ourmethod, and test it for both edge detection techniques in the bridge simulator developed. Case studies are performed with and without traffic.  In our experiments, the Canny edge detector based \ourmethod, offers up to 40\% lower task completion time than CNN, The CNN-based \ourmethod, excels in up to 12\% better damage detection and achieves 1.8 times better rewards.
\end{enumerate}

The remainder of this article is organized as follows. Section II discusses the related work. The system model and MDP formulation are presented in Sections III and IV respectively. The PPO-based SHM algorithm is proposed in Section IV. Section V shows the performance evaluation and simulation environment. Section VI concludes the article with some concluding remarks and future research directions.

\section{Related Work}
This section will provide an in-depth analysis of the DRL for path planning for UAVs, followed by a detailed examination of structural health management within this context. The purpose of this analysis is to highlight the significance of these topics and their potential impact on the field.
\subsection{DRL for UAVs}
In \cite{UAVautonomous} a partially observable Markov decision process (POMDP) is formulated and solved using an online DRL algorithm that is designed based on two strictly proved policy gradient theorems within the actor-critic framework. The method directly maps UAVs' raw sensory measurements into control signals for navigation, unlike the traditional approaches.
In \cite{UAV_sparse},  a sparse reward scheme is used to ensure that the solution is not biased towards potentially suboptimal directions. However, the absence of intermediate rewards makes the agent's learning less efficient as informative states are rarely encountered. To overcome this challenge, a prior policy that may perform poorly is assumed to be available to the learning agent. This policy reshapes the behavior policy used for environmental interaction to guide the agent in exploring the state space.
In \cite{DRL_imagesurvey},  the main techniques of DRL and its applications in image processing are summarized.
In \cite{UAV_dataCollect}, the twin-delayed deep deterministic policy gradient (TD3) is used to design the UAV's trajectory. A TD3-based trajectory design is presented for completion time minimization, where the UAV's trajectory is optimized to efficiently collect data from multiple Internet of Things (IoT) ground nodes. 
In \cite{UAV_obs_avoid},  the concept of partial observability is discussed, and how UAVs can retain relevant information about the environment structure to make better navigation decisions. The Obstacle Avoidance (OA) technique uses recurrent neural networks with temporal attention and provides better results in terms of distance covered without collisions compared to prior works.

\subsection{SHM}

Extensive research has been conducted on the application of machine learning and deep learning techniques to develop an intelligent bridge management system. Several studies, including \cite{mlshm20, mlshm21, mlshm22, deepmlAcousticwave}, demonstrate that supervised learning can effectively predict maintenance schedules using labeled data. Recently, researchers have begun exploring the use of deep reinforcement learning (DRL) techniques for bridge health maintenance. Various approaches utilizing DQN have been proposed, as described in \cite{drl_inspData, DRLopportunistic, DRLinference, DRLdecision_loadaging}, to monitor the health of the bridge structure. Hierarchical coordinated reinforcement learning has been used to optimally maintain multiple components of the bridge structure, as detailed in \cite{DRL_multicomponent_maintenance}. Moreover, a case-study highway bridge portfolio is examined in \cite{du2022parameterized}, where a parameterized DQN is implemented to offer superior sequential maintenance decisions that can better help the bridge structures adapt to long-term deterioration. Additionally, \cite{DRL_sensor, DRL_rail} showcases the use of DRL for sensor-based real-time decision-making and rail renewal-based maintenance planning.

In \cite{uavshm} provides a detailed comparison between traditional machine learning and transfer learning approaches. 
In \cite{DLshmsurvey} provides valuable insights into the procedure and application of vibration-based, vision-based monitoring, as well as some of the recent technologies used for SHM, including UAVs, sensors, and other related tools. The authors discuss the importance of using these technologies for efficient and effective SHM.
In \cite{bridgeUAV} proposes a method for determining the safety risk level of post-earthquake bridge structures. The evaluation indices are established after considering various factors such as damage type, spalling area, width of cracks, and recognition statistics of all images. 
In \cite{UAv_earthquakBridge} explores the potential of UAV technologies in bridge inspection and monitoring. The authors propose a method for quantifying and visualizing the progression of damage for each structural element, which can help in conveniently monitoring the health of a bridge.
In \cite{UAVfeature} presents a ranging method for calculating and quantifying the real bridge crack widths using corresponding object distance data. This approach can help in accurately predicting the potential damage in the bridge and taking necessary preventive measures.

\section{System Model}

\begin{figure}
    \centering
    \includegraphics[width= \linewidth]{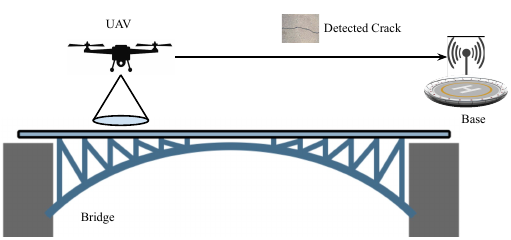}
    \caption{System Model illustrating bridge survey and crack detection using UAV.}
    \label{fig:sysmodel}
\end{figure}

In this work, we consider a bridge as our simulation environment. A bridge is represented as a two-dimensional space where a UAV operates for crack detection during ongoing traffic. An illustration of the system model is presented in Fig. \ref{fig:sysmodel}. The UAV surveys a bridge deck and sends the crack images along with their locations to the base.

\subsection{Simulation Environment}

Our simulation environment mimics the structural layout of a bridge, providing a two-dimensional representation where the UAV navigates for crack detection amidst ongoing traffic. The bridge is conceptualized as a digital space with discrete coordinates, denoted as ($x_t$, $y_t$), which is the position of the UAV at time $t$ relative to the bridge's surface. At a given moment, the UAV maintains a consistent height $h$ above the bridge, ensuring optimal scanning and detection capabilities. To accurately simulate the dynamic conditions encountered during real-world operations, the simulation environment incorporates factors such as varying traffic densities. 

At the beginning of each episode, cracks were located at random locations and the UAV had no prior knowledge of their location. Different types of cracks were simulated, including line cracks, cracks with forks, and curved cracks. A forked crack was created by extending a line crack at a random angle and length to form a fork. A curved crack was created using cubic Bezier curves, which allowed for the creation of parabolic, s-shaped, and other curve-shaped cracks. The length of all cracks was chosen randomly within an acceptable range. The types of cracks are chosen at random when given the number of cracks to be simulated. 

\subsection{Crack Detection Techniques}
The system model is presented in this subsection for bridge inspections and to test our crack detection method. Using deep learning approaches, we aim to improve SHM and quickly address infrastructure damages.

\subsubsection{Canny Edge Detection}
\begin{figure}
    \centering
    \includegraphics[width = \linewidth]{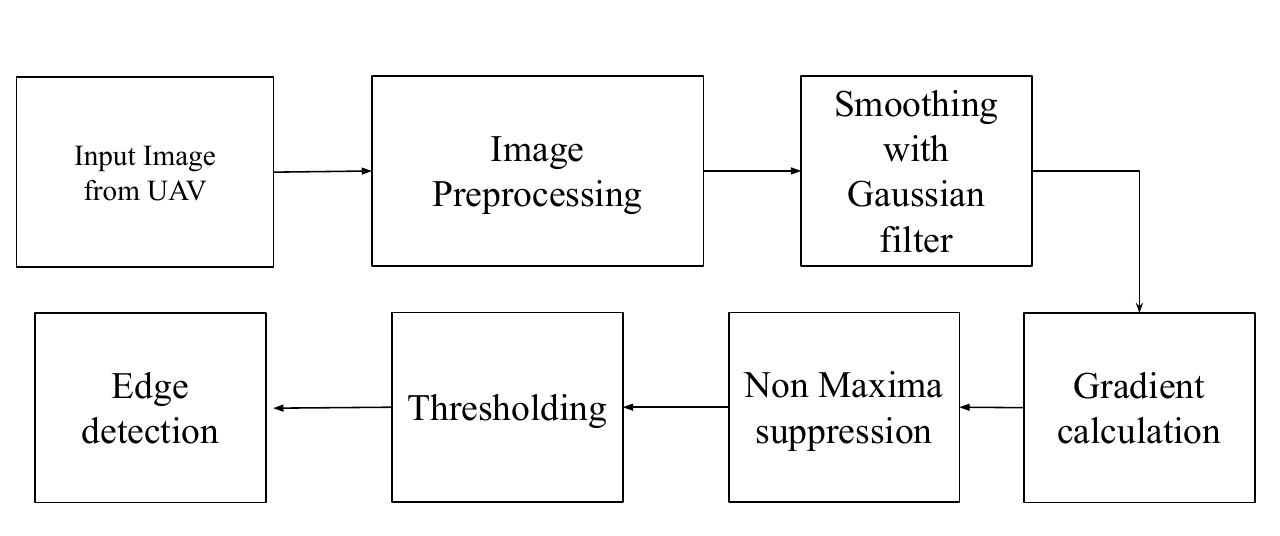}
    \caption{Worflow for Canny Edge Detector technique}
    \label{fig:edge}
\end{figure}

The Canny edge detector is a well-known technique in image processing that is highly effective at accurately identifying object edges within images while reducing noise and false detections. The Canny edge detector is known for its accuracy and robustness, making it widely used in tasks such as object detection, image segmentation, and feature extraction where precise edge detection is essential for analysis and interpretation.  The workflow of the canny edge detector is depicted in Fig. \ref{fig:edge}. The input image from the UAV is fed to the image pre-processing, where the images are converted to grayscale. The method works by first identifying areas of significant intensity changes that could potentially indicate edges. Then a Gaussian filter is applied to smoothen the image. It then uses non-maximum suppression to refine edge detection by retaining only local gradient maxima. Finally, edge tracking by hysteresis or thresholding is utilized to connect strong edge pixels while discarding weak ones. This results in a binary image where edges are highlighted against a black background.

\subsubsection{Transfer Learning based CNN}
\begin{figure}
    \centering
    \includegraphics[width = \linewidth]{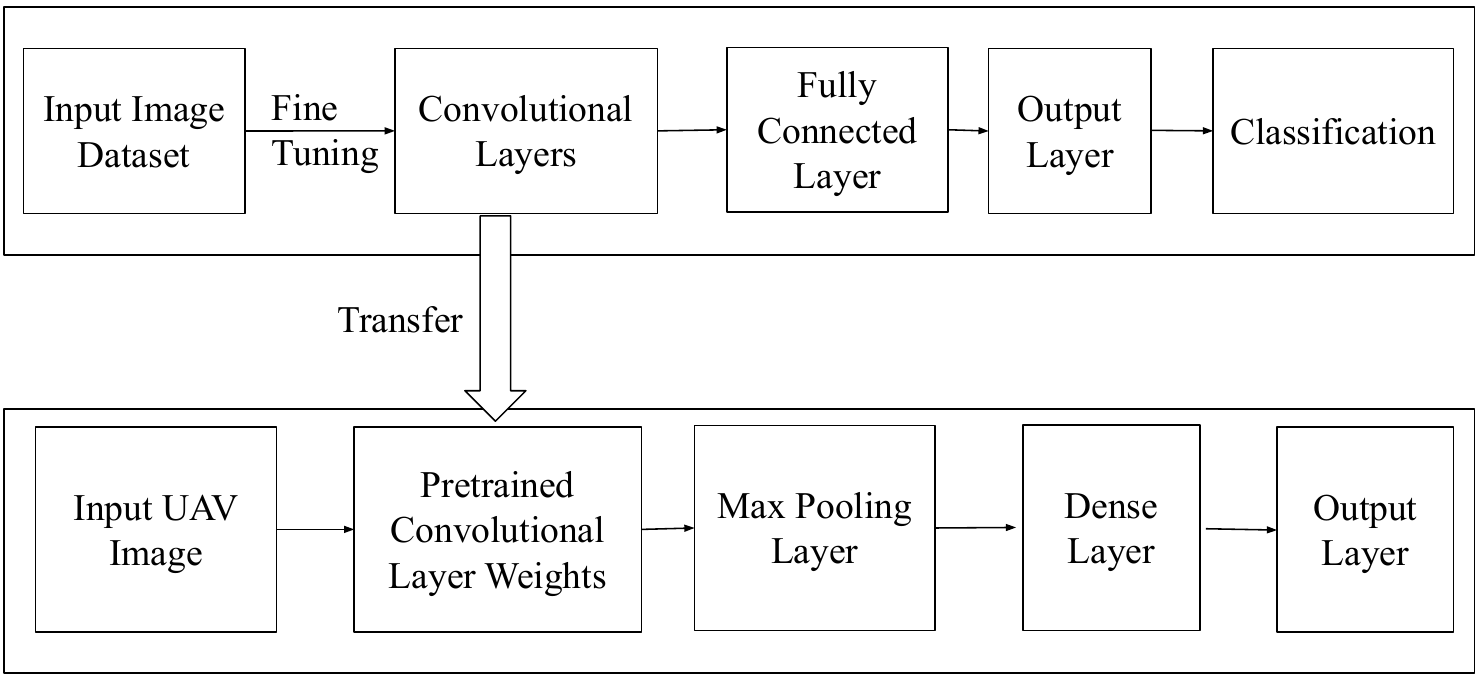}
    \caption{Transfer Learning based CNN Workflow}
    \label{fig:cnn}
\end{figure}

Transfer learning is applied to CNNs to detect concrete cracks by reusing pre-trained CNN models initially trained on large-scale image dataset\footnote{ https://data.mendeley.com/datasets/5y9wdsg2zt/2} and adapting them for the specialized task of crack detection. CNNs can extract meaningful features from images, which can be used for bridge crack detection. The transfer learning workflow is presented in Fig. \ref{fig:cnn}. The convolutional layers of the pre-trained model, which capture hierarchical visual patterns, are retained, while the fully connected layers responsible for image classification are replaced with additional layers specifically designed for crack detection, such as convolutional, pooling, and fully connected layers. During the fine-tuning process, the model learns to recognize crack-related features by adjusting its parameters using a dataset that contains annotated images of concrete surfaces with crack locations. The fine-tuned model is then evaluated on a separate validation dataset to assess its performance in accurately detecting concrete cracks. In this work, we leverage a transfer learning approach with a pre-trained CNN model that achieves a 99.6 percent accuracy when trained and tested on our image dataset. This high level of accuracy underscores the effectiveness of our model in accurately detecting concrete cracks. This approach not only speeds up the training process but also enhances the model's performance and generalization capabilities, making it well-suited for real-world applications in crack detection of bridges.

\section{Markov Decision Process}

A Markov Decision Process (MDP) for crack detection in a SHM context involves defining states, actions, and rewards that are essential to the decision-making process of an agent for inspecting a bridge for cracks. 

\subsection{State}
The problem at hand involves identifying and assessing potential crack locations on a bridge grid. These locations are identified by the coordinates $s_c = (x_c, y_c)$. Along with the potential crack locations, we also need to consider the current location of the UAV on the bridge grid, which is represented by $s_u = (x_u, y_u)$. Moreover, we need to take into account the presence or absence of obstructing traffic. This is represented by a binary variable $s_t$, where a value of 1 indicates the presence of traffic and a value of 0 indicates its absence. The UAV does not know the location of the cracks till it encounters one. All the detected crack locations are stored. Also, all the visited locations are stored in $v_l$. So, the state can be expressed as follows:
\begin{equation}
    S =\{s_c, s_u, s_t, v_l\}.
\end{equation}
If $s_t$ equals 1, it indicates the presence of ongoing traffic on the deck. In such cases, the UAV pauses until the traffic clears up. However, if the UAV pauses for a certain amount of time and the traffic doesn't clear up, it skips the location and revisits it later.

\subsection{Action}
The system has two distinct types of actions that the UAV can take in order to navigate the bridge environment effectively. The first type of action involves the UAV choosing a direction to move in, which is denoted by $a_d$. The available options for $a_d$ include up, down, left, and right, which enable the UAV to move one unit in the corresponding direction. The UAV flies at a constant height. The second type of action involves the UAV pause itself at a specific location on the grid, which is represented by the action $a_p$. This type of action is particularly useful when the UAV needs to pause in a specific location because of traffic. 

\subsection{Reward}
The rewards assigned to the UAV within the MDP framework play a crucial role in guiding its decision-making process during the surveillance task. The system assigns different rewards for various outcomes, which motivate the UAV to complete its mission efficiently while conserving energy. The rewards can be expressed in the following equations:

Temporal reward: The temporal reward is a negative reward incurred by the UAV for every time step, encouraging efficient completion of the surveillance task. The temporal reward can be expressed as,
$ r_m = -1 $.

Pause reward: This is a neutral reward received by the UAV when it does not move while waiting for the traffic to clear up. At time $t$, if an action $a_t$ is the action to pause $a_p$ then, the reward is 0. The pause reward is expressed as, $r_{p} = 0 \ \text{when \ $ a_t =a_p$}$.

Visited reward: This is a penalty imposed when the UAV revisits a location that has already been surveyed, discouraging redundant scanning. The pause reward is expressed as, $ r_v = -1$.

Crack detection reward: This is a positive incentive given to the UAV for detecting new cracks, which encourages thorough surveillance.  The crack detection reward is expressed as, $ r_c = +10$.

New location reward: This is a positive reward earned by the UAV for surveying a previously unexplored location, and encouraging exploration. The new location reward is expressed as, $r_{nl} = +5$.

End of episode reward: This is a high positive reward received by the UAV upon completing the surveillance task, signaling successful task completion. The end-of-episode reward is expressed as, $r_e = +20$
The total reward is the sum of these rewards,
\begin{equation}
    r_t = r_m + r_p + r_v + r_c + r_n + r_e
\end{equation}

\subsection{Environment}
In this framework, the agent is represented by a UAV that operates autonomously to navigate the bridge environment. The UAV is equipped with sensors to scan for cracks and other defects, and it uses the data gathered from these scans to make informed decisions about the best course of action. The decisions made by the UAV are based on the observed states of the bridge and the potential actions available to it. The UAV operates within a state space defined by the bridge's structural characteristics and environmental conditions. We leverage the PPO algorithm, the agent iteratively learns a policy for selecting actions that maximize cumulative rewards while adhering to constraints and objectives specified in the SHM task. Through interactions with the environment, the UAV transitions between states, executes actions, and receives feedback in the form of rewards. This facilitates the optimization of its decision-making process over time. Overall, the agent's objective within the MDP framework is to dynamically navigate the bridge, detect structural damages, and optimize its inspection strategy to ensure effective SHM.

The UAV's location changes based on the selected actions, with new coordinates $s_u' = (x_u', y_u')$ reflecting its updated position on the bridge grid. The presence or absence of obstructing traffic may dynamically transition based on real-time traffic conditions, impacting the UAV's movement options and decision-making process. Throughout the surveillance mission, the total reward $R$ accumulated by the UAV is the sum of all individual rewards received, guiding its behavior and optimizing its performance.

\section{\ourmethod: PPO based SHM}

The agent utilizes available state samples to compute the anticipated reward achievable by executing an action $a$ from a state $s$, constructing a precise estimate of the action-value function $Q^*(s, a)$. Due to the problem's high dimensionality and the complexity of Q-function computation, deep neural networks are employed as function approximators. The PPO agent, employing neural networks as function approximators, adopts a policy gradient approach, seeking to maximize future rewards expressed as the sum of all discounted future returns following time step $t$. The discounted future reward, with the discount factor $\gamma$, is represented as:

\begin{eqnarray}
R_t = r_t + \sum_{{\tau}=0}^{\infty} \gamma^{\tau} r_{t+\tau},
\end{eqnarray}
with $\gamma \in [0,1]$.The policy $\pi(a_t|s_t)$ outlines the agent's actions and reflects the probability of selecting an action $a_t \in A$ based on the observed state $s_t \in S$. The action-value function, under policy $\pi$, represents the expected return of the agent's action in a particular state. It is represented as $Q_{\pi}(s, a) = \displaystyle {\mathbb{E}}{\pi} [R_t |s_t, a_t]$, where $\mathbb{E}{\pi}$ indicates the expected value under adherence to policy $\pi$. The optimal policy $\pi^*$ aligns with the optimal action-value function $Q^*(s, a)= \max_{\pi} Q_{\pi}(s, a)$ and satisfies the Bellman optimality equation. The optimal action-value function is equal to the expected reward of an action $a$, plus the discounted expected value of the best action in the subsequent state $s_{t+1}$ and is expressed as follows,
\begin{equation}
    Q^*(s,a) = \displaystyle \mathop{\mathbb{E}} [R_{t+1}+ \gamma \max_{a'} Q^*(s_{t+1},a_{t+1})| s_t, a_t].
\end{equation}

The highest achievable expected return can be determined by the action-value function $Q^*(s, a)$ under any policy. In order to maximize the Q function, a policy $\pi$ maps states to actions. A new state, $s_{t+1}$ is generated based on the action taken at the previous time step. After the agent decides on an action, the parameters are sent to the UAV. The ultimate objective is to compute the reward in a way that maximizes the long-term return of the entire system.

\begin{figure}
    \centering
    \includegraphics[width = \linewidth]{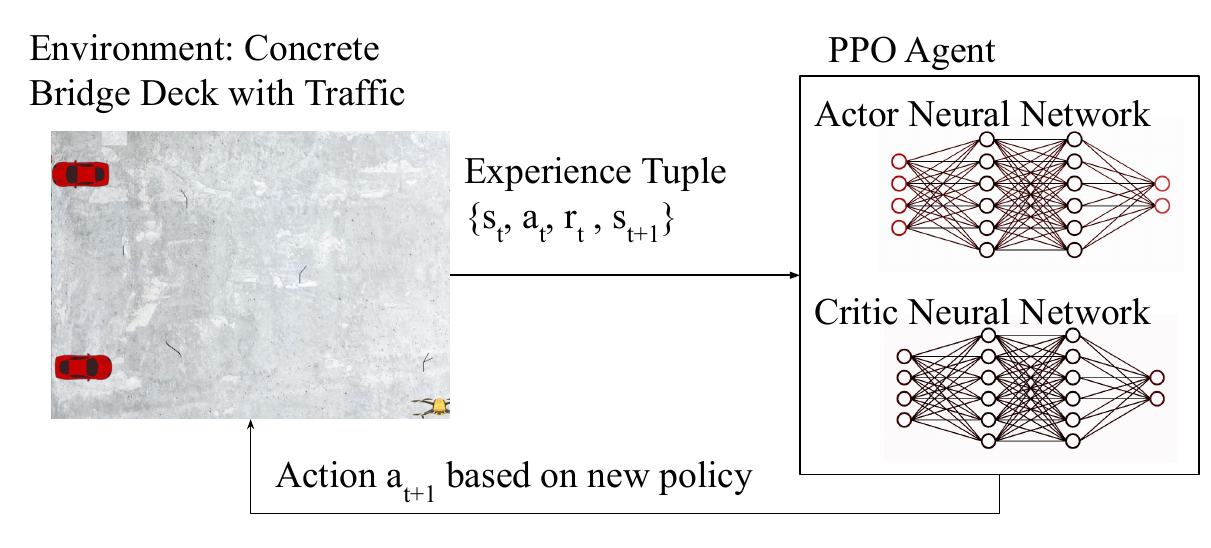}
    \caption{An illustration of the proposed approach with PPO agent and concrete bridge deck with traffic. }
    \label{fig:ppo}
\end{figure}

The PPO algorithm is a recent development in DRL. It is based on the actor-critic method and uses two pivotal components: the actor and critic networks, both implemented as deep neural networks as shown in Fig. \ref{fig:ppo}. The critic network assesses performance, while the actor-network is tasked with policy learning. Both networks contribute to the target network, which is used to generate target values for training the critic. In the training process, experience tuples are stored in replay memory and used for training both the main and target networks. The PPO-based algorithm uses the current state as input to the actor network, which generates a distribution of actions. The agent's action is determined by sampling from this action distribution. The environment provides a reward for the action taken, influencing the critic network. The critic's output becomes a component of the actor's loss function. The actor network generates the policy, and the critic network evaluates the current policy by estimating the advantage function.

\begin{equation} \label{adv}
    \hat{A}_t^{\pi}(s,a) = Q^{\pi}(s,a) - V^{\pi}(s).
\end{equation}
If the advantage of the optimal action in $s$ is equal to zero then the expected return in $s$ is the same as the expected return when being in state $s$ and taking an action $a$. This is because the optimal policy will always choose $a$ in $s$. The advantage of all other actions is negative which means they bring less reward than the optimal action, so they are less advantageous. The adapted surrogate objective, denoted as \( L^{CLIP}(\theta) \), is defined by the following:

\begin{equation}
    L^{CLIP}(\theta) = \hat{\mathbb{E}}_t \left[ \min \left(\rho_t(\theta)\hat{A_t}, \text{clip}\left(r_t(\theta), 1-\delta, 1+\delta\right) \hat{A}_t\right) \right],
\end{equation}
where \( \rho (\theta) = \frac{\pi_{\theta}(a_t|s_t)}{\pi_{\theta_{\text{old}}}(a_t|s_t)} \), \( \rho (\theta_{\text{old}})= 1 \) represents the probability ratio, \( \delta \) is a hyperparameter, and \( \theta \) is the policy parameter. Here, \( \pi_{\theta} \) is a stochastic policy, \( \theta_{\text{old}} \) is the vector of policy parameters before an update, and \( \rho (\theta) \) measures the ratio between the action under the current policy and the action under the previous policy. The second part of the equation, \( \text{clip}\left(r_t(\theta), 1-\delta, 1+\delta\right) \hat{A}_t \), adjusts the objective by constraining the probability ratio within the interval \( [1-\delta, 1+\delta] \), mitigating the incentive for \( \rho_t \) to deviate outside this range. The objective is further refined by taking the minimum of the clipped and unclipped terms, establishing a lower bound on the unclipped objective. This constraint is introduced to prevent excessive policy updates that could result from unconstrained maximization of \( L \). The overall objective is defined as:

\begin{equation}
    L_t(\theta) = \hat{\mathbb{E}}_t \left[L^{CLIP}_t(\theta) - c_1 L_t^{VF} + c_2 H[\pi_{\theta}|s_t]\right],
\end{equation}

where \( L_t^{VF} = (V_{\theta}(s_t) - V_t^{\text{targ}})^2 \) is a value-function error term, \( c_1 \) is a positive constant scaling this term, and \( V_t^{\text{targ}} \) denotes the state-value function of the training target. The entropy of the policy \( \pi \), denoted as \( H[\pi_{\theta}|s_t] \), scaled by the positive constant \( c_2 \), is included in the objective to encourage exploration. The training process of the overall workflow is outlined in Algorithm \ref{alg:PPO}, employing a feed-forward network with two hidden layers as the policy network. Sampling is performed from a single environment, and the default trajectory length is set to 2048.
 Algorithm \ref{alg:PPO} begins by initializing the policy and value function parameters along with hyperparameters such as the learning rate and discount factor. In each episode, the UAV follows its policy for a set number of time steps, collecting trajectories and computing advantages based on the observed rewards. A variable $t_{pause}=0$ is incremented every time the UAV pauses. The UAV cannot pause for more than $\delta$ timesteps. In our implementation $\delta$ is set to 200. During each time step, the UAV selects actions based on the current policy, computes rewards, and collects transitions representing state-action-reward-next state sequences. The policy and value function parameters are updated periodically using policy gradients, with gradient clipping to prevent large updates. This iterative process continues for the specified number of episodes, allowing the UAV to learn an optimal policy for navigating the environment and maximizing cumulative rewards. Through this algorithm, the UAV can autonomously make decisions while effectively adapting its behavior based on the feedback received from the environment.

\begin{algorithm}
\caption{\ourmethod }\label{alg:PPO}
\begin{algorithmic}[1]
\State \textbf{Initialize} policy parameters $\theta_0$, value function parameters $\phi_0$, and clipping threshold $\delta$
\State \textbf{Hyperparameters:} learning rate $\alpha$, discount factor $\gamma$, clipping parameter $\epsilon$
\For {episodes $k=1, \dots, E$}
    \State Follow policy, $\pi_{\theta_{\text{old}}}$ for $\tau$ time steps 
    \State Collect trajectories: $\mathbb{D} = \tau_i$ set of trajectories
    \State Compute advantages: $A(s_t, a_t) = Q(s_t, a_t) - V_\phi(s_t)$
    \State  $t_{pause}=0$
    \For {epoch $t=1, \dots, \tau$}
        
        \If{No Traffic or $t_{pause} \geq \delta $} 
        \State Take action $a_t$ to move
        \ElsIf{Traffic presence and $t_{pause}< \delta$} 
        \State Take action $a_p$ to pause
        \State $t_{pause}=t_{pause}+1$
        \EndIf  \textbf{End if}
        \State Collect transition: $(s_t, a_t, r_t, s_{t+1})$
        \If{$t$ is a multiple of $K$}
            \State Compute policy gradient:
             $$\nabla_\theta L_t(\theta) = \frac{1}{N} \sum_{t=1}^N \nabla_\theta \log \pi_\theta(a_t|s_t) A(s_t, a_t)$$
            \State Update policy: $\theta \leftarrow \theta + \alpha \nabla_\theta L_t(\theta)$
            \State Update value function:
            \State $$\phi \leftarrow \phi + \alpha \nabla_\phi \frac{1}{N} \sum_{t=1}^N (V_\phi(s_t) - Q(s_t, a_t))^2$$
            \State Clip policy gradient:
             $$\nabla_\theta L_t(\theta) \leftarrow \text{clip}(\nabla_\theta L_t(\theta), -\epsilon, \epsilon)$$ 
        \EndIf  \textbf{End if}
    \EndFor  \textbf{End for}
    \State $\theta_{\text{old}} \leftarrow \theta$
\EndFor  \textbf{End for}
\end{algorithmic}
\end{algorithm}

\section{Results}

In this section, the convergence analysis of the proposed \ourmethod\  algorithms is presented. The edge detector-based \ourmethod\ is analyzed in various scenarios with respect to the CNN-based \ourmethod. We compare the proposed approach with a random policy approach. In the random policy implementation, the UAV can take random actions within the simulator using the edge detection technique. Then a summary of the analysis is provided based on the impact of several hyper-parameters on both \ourmethod\ algorithms.

\subsection{Implementation details}
We have defined a bridge area of 800 m × 600 m where the UAV can fly at 50 m altitude. By default, there is 1 UAV at the base, along with 5 cracks placed at unknown locations on the bridge and 2 cars. Unless stated otherwise, this scenario is considered the default setting. Please see the specific parameters of the environment in Table~\ref{tab:env_params}.

The proposed \ourmethod\ algorithms are implemented on a workstation equipped with 2 GPUs of 32 GB memory. The software environment is configured with \texttt{Python 3.8.12}, \texttt{Tensorflow 2.11.0}, \texttt{stable-baselines 2.10.2}, and \texttt{gym 0.26.2} packages. Additionally, the required software dependencies include \texttt{numpy 1.21.6}, \texttt{pygame 2.5.2}, \texttt{keras 2.11.0}, \texttt{opencv-python 4.8.0.76}, and \texttt{matplotlib 3.5.3}. The hyperparameters utilized in the implementation are detailed in \tablename~\ref{tab:ppo_hprams}.


\begin{table}[]
\centering
\caption{Environmental Parameters}
\label{tab:env_params}
\begin{tabular}{|l|l|}
\hline
\textbf{ Parameters }    & \textbf{Value }           \\ \hline
Length                                 & 800 m                                     \\ \hline
Breadth                             & 600 m                                     \\ \hline
Height of UAV            & 50 m                                       \\ \hline
Maximum crack width                                 & 1 m                                      \\ \hline
 Maximum crack height             & 20 m                                       \\ \hline
Scanning range of UAV              & 100 m                                      \\ \hline
Maximum speed of the UAV         & 25 m/s                                     \\ \hline
\end{tabular}
\end{table}

\begin{table}
\centering
\caption{PPO Hyper-parameters}
\label{tab:ppo_hprams}
\begin{tabular}{|l|l|}
\hline
\textbf{ Parameters }    & \textbf{Value }           \\ \hline
Neurons in hidden layer 1    & $256$                                                \\ \hline
Neurons in hidden layer 2    & $256$                                                \\ \hline
Relay memory size            & $3072$                                               \\ \hline
Minibatch size               & $768$                                                \\ \hline
Learning rate             & $1\times10^{-4}$ 
     \\ \hline
Discount factor         & $0.99$                                               \\ \hline
Activation function               & ReLU                                                 \\ \hline
Clip range                       & $0.2$                                                \\ \hline
Optimizer                         & Adam                                                 \\ \hline
Epochs                           & $20$                                                 \\ \hline
Total episodes                 & $5\times 10^{5}$                                     \\ \hline
\end{tabular}
\end{table}

\begin{figure*}[htbp]
  \centering
  \begin{minipage}[b]{0.45\linewidth}
    \centering
    \includegraphics[width=\linewidth]{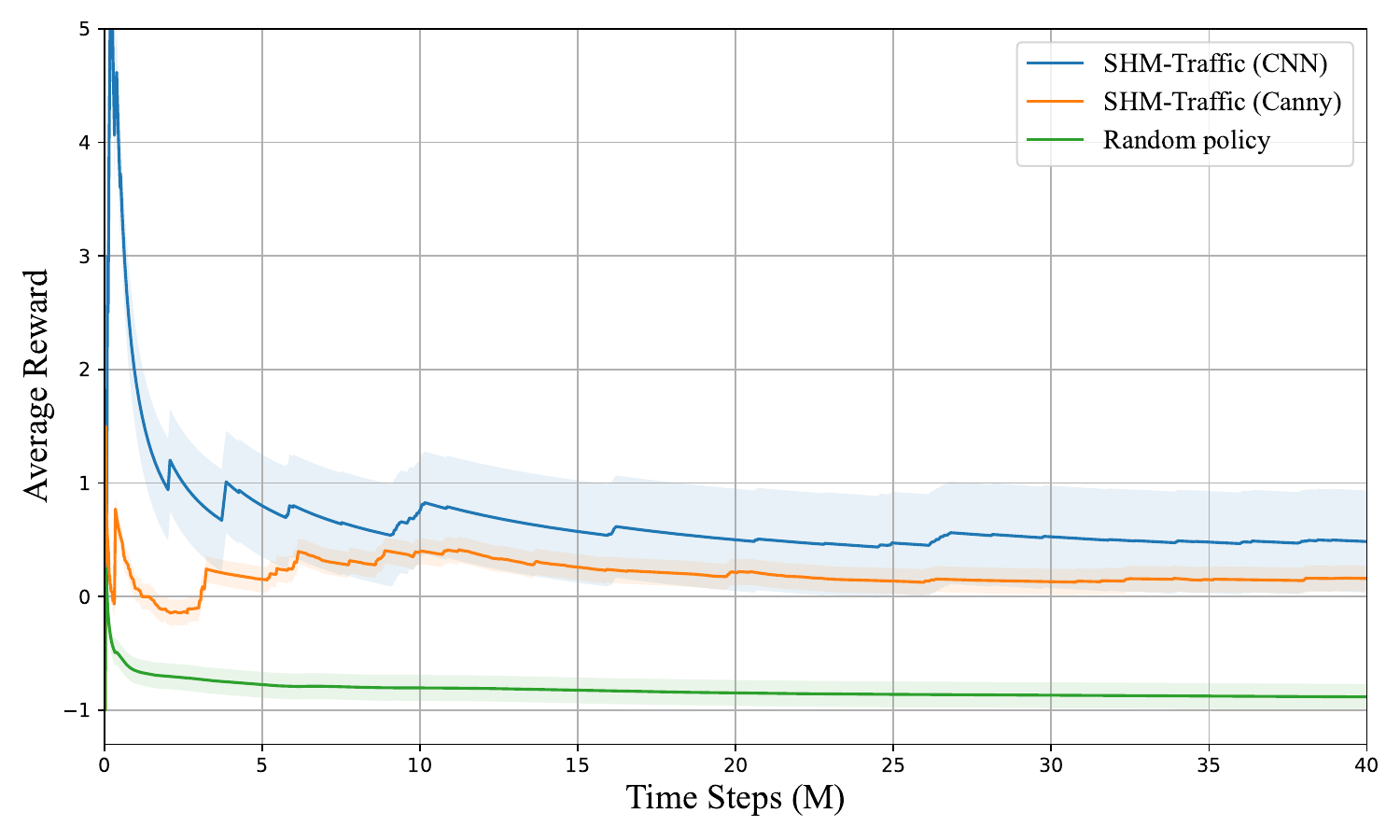}
    \caption{PPO performance comparison for 5 cracks with canny edge detector versus CNN. CNN outperforms the Canny edge detector.}
    \label{fig:5cr}
  \end{minipage}
  \hfill
  \begin{minipage}[b]{0.45\linewidth}
    \centering
    \includegraphics[width=\linewidth]{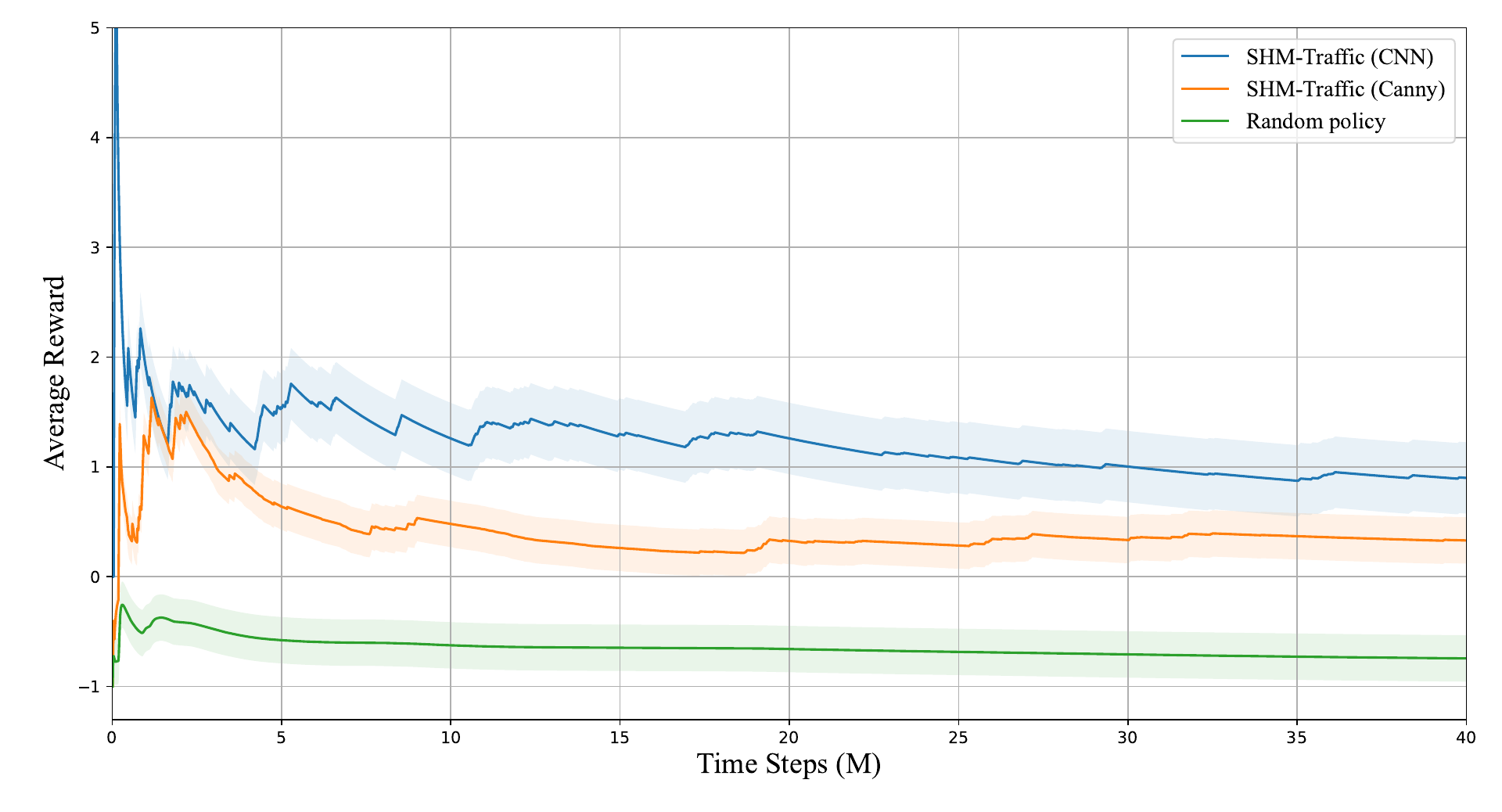}
     \caption{{PPO performance comparison for 10 cracks with canny edge detector, CNN, and random policy. CNN outperforms both the Canny edge detector and random policy.}}
    \label{fig:10cr}
  \end{minipage}

  \begin{minipage}[b]{0.45\linewidth}
    \centering
    \includegraphics[width=\linewidth]{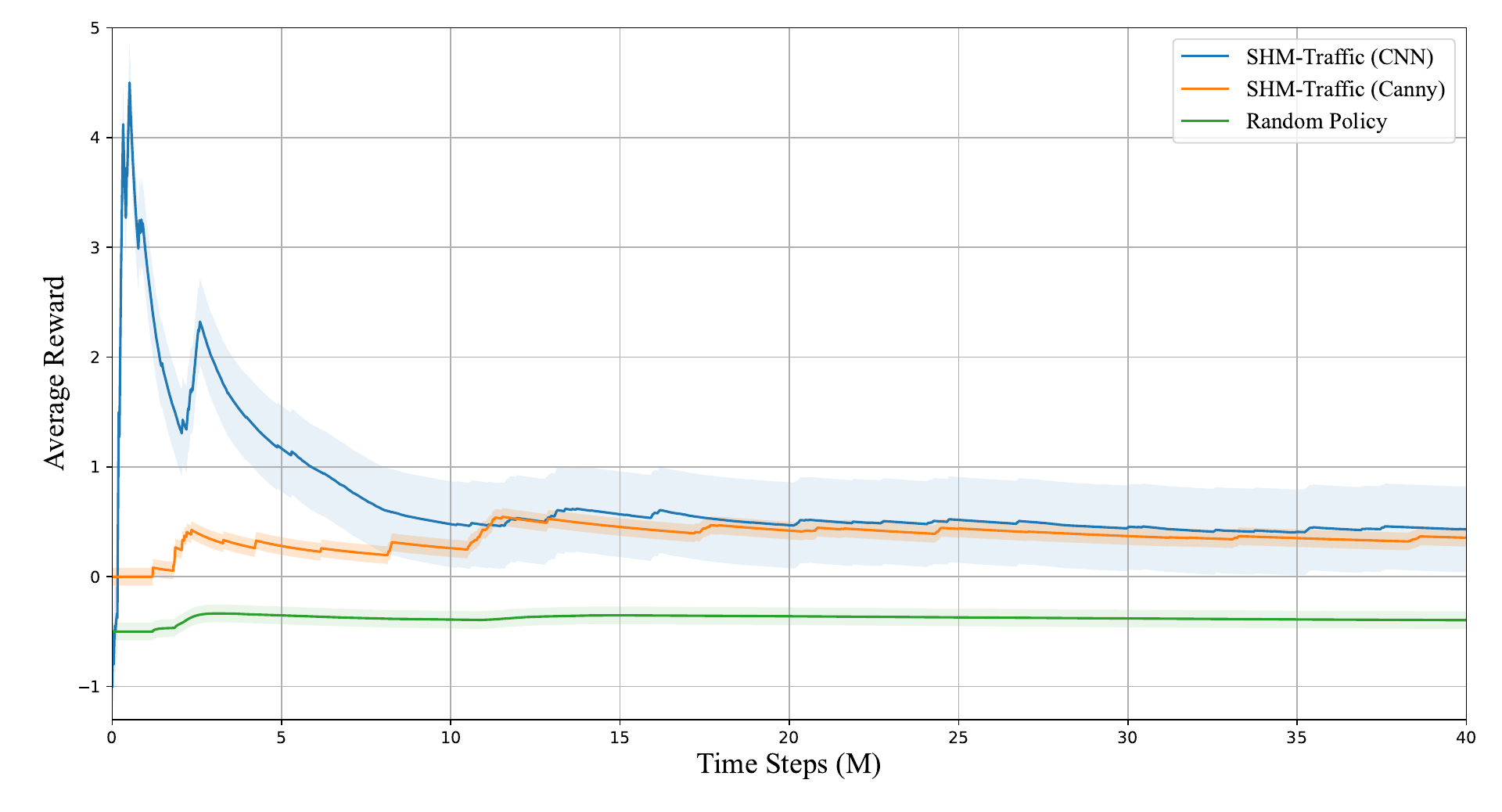}
    \caption{{PPO performance comparison for 5 cracks and 2 cars with canny edge detector, CNN, and random policy. CNN outperforms the Canny edge detector, which in turn outperforms random policy.}}
    \label{fig:5cr2ca}
  \end{minipage}
  \hfill
  \begin{minipage}[b]{0.45\linewidth}
    \centering
    \includegraphics[width=\linewidth]{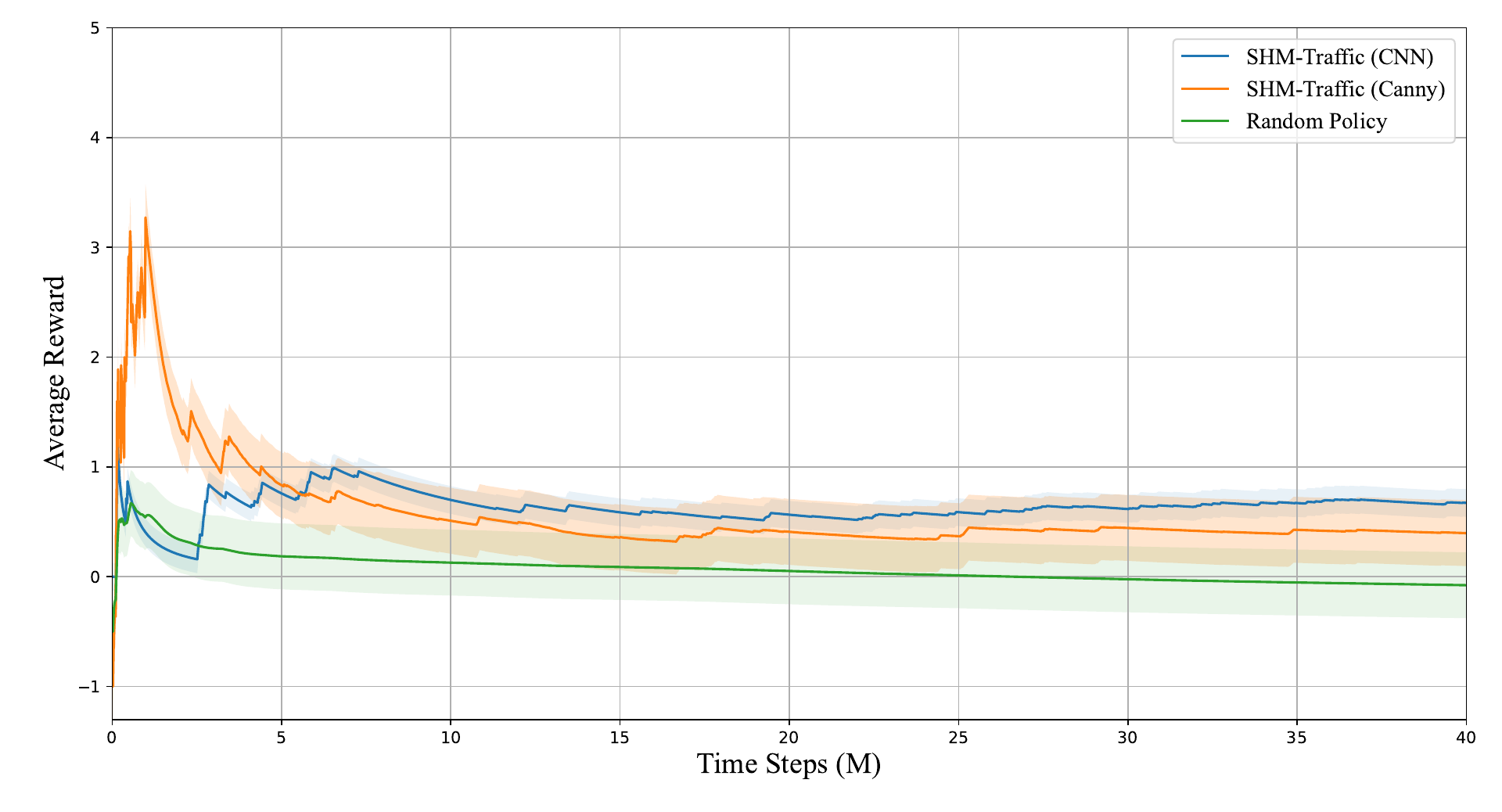}
    \caption{{PPO performance comparison for 10 cracks and 2cars with canny, CNN, and random policy. CNN outperforms the Canny edge detector and random policy.}}
    \label{fig:10cr2ca}
    \label{fig:comp_cpu}
  \end{minipage}

\end{figure*}
\subsection{Crack detection Methods}

\begin{table}
\scriptsize
    \centering
    \begin{tabular}{|c|c|c|c|}
\hline
Performance  & Average Runtime & Training time & Accuracy \\
\hline
Canny Edge Detector     & 22 ms & 20 epochs & 87\%\\
\hline
  CNN   & 60 ms & 20 epochs & 99.6\%\\
  \hline
    \end{tabular}
    \caption{Performance Comparison of Canny edge detector VS CNN for crack detection}
    \label{tab:comp}
\end{table}

The Canny edge detector and CNN represent two distinct approaches to crack detection tasks in SHM systems for bridges, each with its own advantages and considerations. A brief description of the canny edge detector and the pre-trained convolutional layer is presented here as image classification is not the primary focus of this work. We focus on leveraging these crack detection methods in bridge surveys with traffic using DRL. The canny edge detector method was implemented with edge evaluation threshold $\geq 0.6.$ The Canny edge detector stands out as a computationally efficient option, boasting an average runtime of 22ms. In contrast, the CNN exhibits a slightly longer average runtime of 60ms, primarily due to the computational overhead associated with deep learning-based inference processes. The CNN model was trained on 32000 images and validated on 8000 images. The maximum accuracy achieved was 99.6\%. The pre-trained CNN weights from \cite{cnn} are used in this work. The trained CNN has 6 convolutional layers. Additionally, the CNN necessitates training time to learn and optimize its parameters for crack detection, a process we conducted over 20 epochs in our experiments to achieve the desired level of performance. Despite the longer runtime and training requirements, the CNN offers distinct advantages over the Canny edge detector. The CNN's superior accuracy and adaptability through training make it a compelling option for applications where precision and flexibility are paramount. By learning from data, CNN can effectively discern intricate patterns and variations in cracks, potentially enhancing detection performance in complex scenarios. In contrast, the Canny edge detector may struggle with noisy backgrounds. The choice between these methods depends on the specific requirements of the application. A comparison is presented in Table \ref{tab:comp}. When runtime efficiency is critical, the Canny edge detector may be preferred, especially for real-time monitoring applications where quick processing is essential. However, in situations where detection accuracy and adaptability are paramount, the CNN emerges as a more suitable choice despite its longer runtime and training overhead. This comparative analysis aids decision-making in selecting the most appropriate crack detection method tailored to the needs of SHM systems for bridges, striking a balance between runtime efficiency and detection accuracy.



\subsection{Performance with no traffic}

In this scenario, we evaluate the performance of \ourmethod\ under ideal conditions with no obstructing traffic on the bridge. Fig. \ref{fig:5cr} illustrates the comparison of our method's performance for detecting 5 cracks using both the Canny edge detector and CNN. Similarly, Fig. \ref{fig:10cr} presents the comparison for detecting 10 cracks. The initial high reward in \ourmethod (CNN) is because the agent finds the crack early in the experiment, which results in a high positive reward. As time goes by the high positive rewards for crack detection are compensated by the negative rewards such as temporal rewards and visited rewards. In the case of \ourmethod (Canny), the agent doesn't immediately find the crack and hence there is no spike in the initial time steps. This behavior depends on the location of the crack and the random actions taken by the UAV. However, in our observation, the random policy-based UAV action revisits the visited locations many times resulting in a lower average reward. In analyzing the results across both scenarios, we consistently observe that the CNN outperforms both the Canny edge detector and random policy in terms of both crack detection accuracy and task completion time. This implies that the utilization of deep learning techniques, particularly CNN, yields significant improvements in crack detection during bridge inspection tasks conducted under optimal conditions. The superiority of the CNN can be attributed to its ability to learn intricate patterns and features inherent in crack formations, allowing for more precise and accurate detection compared to traditional image processing methods like the Canny edge detector. Additionally, CNN's adaptability and generalization capabilities contribute to its effectiveness across different crack detection scenarios.

\subsection{Performance with traffic}

To assess the robustness of our method in real-world scenarios, we introduce obstructing traffic on the bridge and evaluate its performance under these conditions. This scenario simulates the presence of external disturbances that can affect the UAV's navigation and crack detection capabilities. Fig. \ref{fig:5cr2ca} and Fig. \ref{fig:10cr2ca} depict the reward convergence of \ourmethod\ for detecting 5 and 10 cracks, respectively, in the presence of two cars on the bridge. Despite the presence of traffic disturbances, we consistently observe that the CNN maintains superior performance compared to the Canny edge detector in terms of crack detection accuracy and task completion time.  Both CNN and Canny edge detector show better performance than the random policy approach. This reaffirms the effectiveness of CNN in challenging environments with traffic disturbances, highlighting its robustness and adaptability in real-world scenarios.



\begin{figure}
    \centering
    \includegraphics[width= 0.9\linewidth]{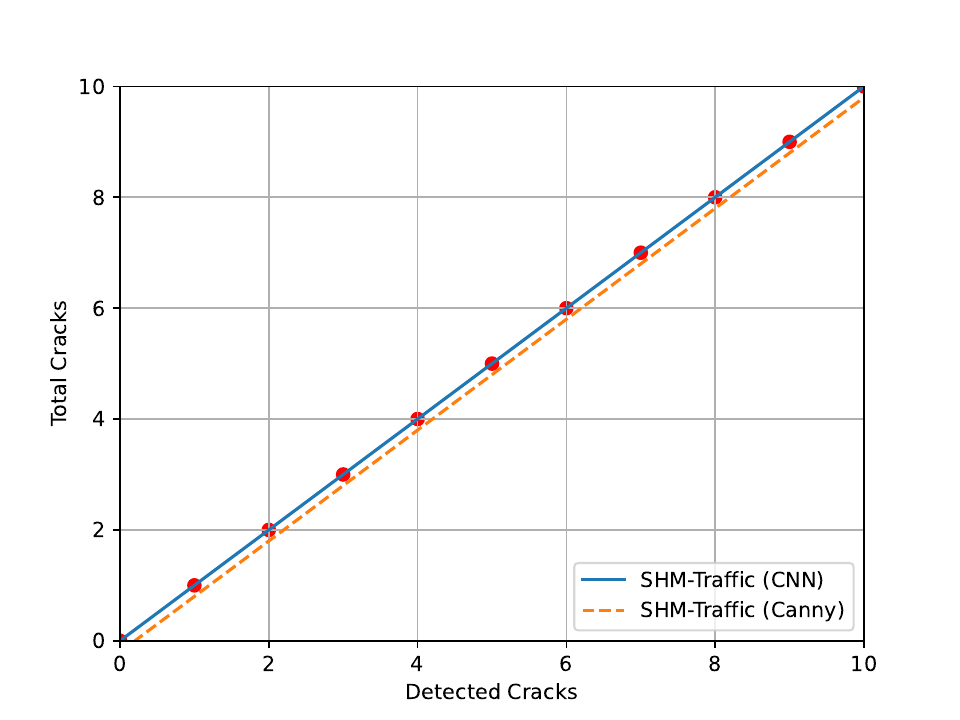}
    \caption{A trend showing total cracks versus detected cracks in both schemes. The UAV can successfully detect all the cracks but the time taken to finish the task differs.}
    \label{fig:detected}
\end{figure}

\subsection{Performance Comparison for detected cracks and task completion time}
In this subsection, we provide a comprehensive analysis of the detected cracks and task completion time achieved by \ourmethod, aiming to provide a detailed understanding of our approach's performance. Fig. \ref{fig:detected} presents a comparison of the number of detected cracks and the corresponding task completion time for the Canny edge detector, CNN, and random policy approach.  We observe a clear trend where the CNN exhibits higher crack detection rates and achieves task completion in less time compared to the Canny edge detector.

Table \ref{tab:perfcan} summarizes the performance of \ourmethod\ (Canny) when false cracks are simulated on the bridge deck. We observe the average rewards and task completion time. The task completion time is the total time taken by the UAV to finish the bridge survey. We compare different task completion times by taking 10 cracks, no cars, and no false cracks as a reference. We summarize the relative improvement (\%) with respect to the reference. We observe that as the number of cars and the number of false cracks increase, task completion increases. However, the increase in the task completion time is within a tolerable limit. A similar observation is made for \ourmethod\ (CNN) and the observations are summarized in Table \ref{tab:perfcnn}.
The higher crack detection rates observed with the CNN can be attributed to its ability to learn intricate patterns and features inherent in crack formations, enabling more precise and accurate detection. 
Furthermore, CNN's efficiency in completing tasks within a shorter time frame highlights its effectiveness in streamlining inspection processes and minimizing operational downtime. This can have significant implications for bridge maintenance and safety management, as timely detection of structural defects is crucial for preventing potential hazards and ensuring the integrity of infrastructure. Overall, our results underscore the efficacy of leveraging advanced techniques such as deep learning, particularly CNN, in enhancing crack detection accuracy and efficiency in bridge inspection tasks, even in the presence of challenging environmental conditions such as traffic disturbances. These findings have significant implications for improving the reliability and effectiveness of SHM systems for bridges.

\begin{table}
\scriptsize
    \centering
    \begin{tabular}{|c|c|c|c|}
\hline
False cracks & Cars & Average Reward & Task completion time (relative\%)\\
\hline
0 & 0 & 0.36 &  100\\
0 & 2 & 0.4 & 93\\
\hline
1   & 0 & 0.27 & 94.5\\
1   & 2 & 0.2 & 92\\
  \hline
2  & 0 & 0.31 &  94\\
2  & 2 & 0.29 &  89\\
\hline
5  & 0 & 0.19 &  91.1\\
5  & 2 & 0.1 & 88\\
\hline
    \end{tabular}
    \caption{Performance Comparison of \ourmethod (Canny) with 10 real cracks and different false cracks}
    \label{tab:perfcan}
\end{table}

\begin{table}
\scriptsize
    \centering
    \begin{tabular}{|c|c|c|c|}
\hline
False cracks & Cars & Average Reward & Task completion time (relative\%)\\
\hline
0 & 0 & 1.46 &  100\\
0 & 2 & 1.4 & 95.6\\
\hline
1   & 0 & 1.28 & 98\\
1   & 2 & 1.21 & 95\\
  \hline
2  & 0 & 0.89 &  95\\
2  & 2 & 0.5 &  90.3\\
\hline
5  & 0 & 0.53 &  89.6\\
5  & 2 & 0.41 & 84.5\\
\hline
    \end{tabular}
    \caption{Performance Comparison of \ourmethod (CNN) with 10 real cracks and different false cracks}
    \label{tab:perfcnn}
\end{table}


\section{Conclusion}

This work proposes \ourmethod\ which highlights the effectiveness of integrating DRL approaches, especially with CNN, into SHM processes for bridges. We conducted comprehensive evaluations under various scenarios, including both ideal conditions and realistic settings with obstructing traffic, and demonstrated the superior performance of \ourmethod\ based on CNN over other methods such as the \ourmethod\ based on Canny edge detector and random policy. \ourmethod\ (CNN) consistently exhibited higher crack detection rates and completed tasks in less time, accurately identifying structural defects and streamlining inspection processes. This emphasizes the potential of deep learning-based approaches in enhancing the accuracy and reliability of SHM systems for bridges, leading to improved infrastructure integrity and enhanced public safety. Further research and development in this field holds promise for advancing the state-of-the-art in bridge monitoring and maintenance practices, ultimately contributing to the longevity and resilience of critical transportation infrastructure worldwide.
In the future, we plan to incorporate multiple UAVs and create a digital twin of the surveyed bridge using the same methodology proposed in this work.
\bibliographystyle{IEEEtran}
\bibliography{IEEEabrv,reference.bib}

\begin{thebibliography}{10}
\providecommand{\url}[1]{#1}
\csname url@samestyle\endcsname
\providecommand{\newblock}{\relax}
\providecommand{\bibinfo}[2]{#2}
\providecommand{\BIBentrySTDinterwordspacing}{\spaceskip=0pt\relax}
\providecommand{\BIBentryALTinterwordstretchfactor}{4}
\providecommand{\BIBentryALTinterwordspacing}{\spaceskip=\fontdimen2\font plus
\BIBentryALTinterwordstretchfactor\fontdimen3\font minus \fontdimen4\font\relax}
\providecommand{\BIBforeignlanguage}[2]{{%
\expandafter\ifx\csname l@#1\endcsname\relax
\typeout{** WARNING: IEEEtran.bst: No hyphenation pattern has been}%
\typeout{** loaded for the language `#1'. Using the pattern for}%
\typeout{** the default language instead.}%
\else
\language=\csname l@#1\endcsname
\fi
#2}}
\providecommand{\BIBdecl}{\relax}
\BIBdecl

\bibitem{mlshm20}
F.-G. Yuan, S.~A. Zargar, Q.~Chen, and S.~Wang, ``Machine learning for structural health monitoring: challenges and opportunities,'' \emph{Sensors and smart structures technologies for civil, mechanical, and aerospace systems 2020}, vol. 11379, p. 1137903, 2020.

\bibitem{mlshm21}
M.~Flah, I.~Nunez, W.~Ben~Chaabene, and M.~L. Nehdi, ``Machine learning algorithms in civil structural health monitoring: A systematic review,'' \emph{Archives of computational methods in engineering}, vol.~28, pp. 2621--2643, 2021.

\bibitem{uavshm}
H.~Malik, A.~Alzarrad, and E.~Shakshuki, ``Payload assisted unmanned aerial vehicle structural health monitoring (uavshm) for active damage detection,'' \emph{Procedia Computer Science}, vol. 210, pp. 78--85, 2022.

\bibitem{mlshm22}
A.~Malekloo, E.~Ozer, M.~AlHamaydeh, and M.~Girolami, ``Machine learning and structural health monitoring overview with emerging technology and high-dimensional data source highlights,'' \emph{Structural Health Monitoring}, vol.~21, no.~4, pp. 1906--1955, 2022.

\bibitem{mnih2015human}
V.~Mnih, K.~Kavukcuoglu, D.~Silver, A.~A. Rusu, J.~Veness, M.~G. Bellemare, A.~Graves, M.~Riedmiller, A.~K. Fidjeland, G.~Ostrovski \emph{et~al.}, ``Human-level control through deep reinforcement learning,'' \emph{nature}, vol. 518, no. 7540, pp. 529--533, 2015.

\bibitem{zhao2021structural}
S.~Zhao, F.~Kang, J.~Li, and C.~Ma, ``Structural health monitoring and inspection of dams based on uav photogrammetry with image 3d reconstruction,'' \emph{Automation in Construction}, vol. 130, p. 103832, 2021.

\bibitem{edgedet}
Y.-T. Zhou, V.~Venkateswar, and R.~Chellappa, ``Edge detection and linear feature extraction using a 2-d random field model,'' \emph{IEEE Transactions on Pattern Analysis and Machine Intelligence}, vol.~11, no.~1, pp. 84--95, 1989.

\bibitem{zhu2023transfer}
Z.~Zhu, K.~Lin, A.~K. Jain, and J.~Zhou, ``Transfer learning in deep reinforcement learning: A survey,'' \emph{IEEE Transactions on Pattern Analysis and Machine Intelligence}, 2023.

\bibitem{schulman2017proximal}
J.~Schulman, F.~Wolski, P.~Dhariwal, A.~Radford, and O.~Klimov, ``Proximal policy optimization algorithms,'' \emph{arXiv preprint arXiv:1707.06347}, 2017.

\bibitem{UAVautonomous}
C.~Wang, J.~Wang, Y.~Shen, and X.~Zhang, ``Autonomous navigation of uavs in large-scale complex environments: A deep reinforcement learning approach,'' \emph{IEEE Transactions on Vehicular Technology}, vol.~68, no.~3, pp. 2124--2136, 2019.

\bibitem{UAV_sparse}
C.~Wang, J.~Wang, J.~Wang, and X.~Zhang, ``Deep-reinforcement-learning-based autonomous uav navigation with sparse rewards,'' \emph{IEEE Internet of Things Journal}, vol.~7, no.~7, pp. 6180--6190, 2020.

\bibitem{DRL_imagesurvey}
W.~Fang, L.~Pang, and W.~Yi, ``Survey on the application of deep reinforcement learning in image processing,'' \emph{Journal on Artificial Intelligence}, vol.~2, no.~1, pp. 39--58, 2020.

\bibitem{UAV_dataCollect}
Y.~Wang, Z.~Gao, J.~Zhang, X.~Cao, D.~Zheng, Y.~Gao, D.~W.~K. Ng, and M.~Di~Renzo, ``Trajectory design for uav-based internet of things data collection: A deep reinforcement learning approach,'' \emph{IEEE Internet of Things Journal}, vol.~9, no.~5, pp. 3899--3912, 2021.

\bibitem{UAV_obs_avoid}
A.~Singla, S.~Padakandla, and S.~Bhatnagar, ``Memory-based deep reinforcement learning for obstacle avoidance in uav with limited environment knowledge,'' \emph{IEEE transactions on intelligent transportation systems}, vol.~22, no.~1, pp. 107--118, 2019.

\bibitem{deepmlAcousticwave}
M.~A. Haile, E.~Zhu, C.~Hsu, and N.~Bradley, ``Deep machine learning for detection of acoustic wave reflections,'' \emph{Structural Health Monitoring}, vol.~19, no.~5, pp. 1340--1350, 2020.

\bibitem{drl_inspData}
X.~Lei, Y.~Xia, L.~Deng, and L.~Sun, ``A deep reinforcement learning framework for life-cycle maintenance planning of regional deteriorating bridges using inspection data,'' \emph{Structural and Multidisciplinary Optimization}, vol.~65, no.~5, p. 149, 2022.

\bibitem{DRLopportunistic}
A.~Valet, T.~Altenm{\"u}ller, B.~Waschneck, M.~C. May, A.~Kuhnle, and G.~Lanza, ``Opportunistic maintenance scheduling with deep reinforcement learning,'' \emph{Journal of Manufacturing Systems}, vol.~64, pp. 518--534, 2022.

\bibitem{DRLinference}
P.~G. Morato, C.~P. Andriotis, K.~G. Papakonstantinou, and P.~Rigo, ``Inference and dynamic decision-making for deteriorating systems with probabilistic dependencies through bayesian networks and deep reinforcement learning,'' \emph{Reliability Engineering \& System Safety}, vol. 235, p. 109144, 2023.

\bibitem{DRLdecision_loadaging}
M.~Cheng and D.~M. Frangopol, ``A decision-making framework for load rating planning of aging bridges using deep reinforcement learning,'' \emph{Journal of Computing in Civil Engineering}, vol.~35, no.~6, p. 04021024, 2021.

\bibitem{DRL_multicomponent_maintenance}
Y.~Zhou, B.~Li, and T.~R. Lin, ``Maintenance optimisation of multicomponent systems using hierarchical coordinated reinforcement learning,'' \emph{Reliability Engineering \& System Safety}, vol. 217, p. 108078, 2022.

\bibitem{du2022parameterized}
A.~Du and A.~Ghavidel, ``Parameterized deep reinforcement learning-enabled maintenance decision-support and life-cycle risk assessment for highway bridge portfolios,'' \emph{Structural Safety}, vol.~97, p. 102221, 2022.

\bibitem{DRL_sensor}
E.~Skordilis and R.~Moghaddass, ``A deep reinforcement learning approach for real-time sensor-driven decision making and predictive analytics,'' \emph{Computers \& Industrial Engineering}, vol. 147, p. 106600, 2020.

\bibitem{DRL_rail}
R.~Mohammadi and Q.~He, ``A deep reinforcement learning approach for rail renewal and maintenance planning,'' \emph{Reliability Engineering \& System Safety}, vol. 225, p. 108615, 2022.

\bibitem{DLshmsurvey}
M.~Azimi, A.~D. Eslamlou, and G.~Pekcan, ``Data-driven structural health monitoring and damage detection through deep learning: State-of-the-art review,'' \emph{Sensors}, vol.~20, no.~10, p. 2778, 2020.

\bibitem{bridgeUAV}
B.~J. Perry, Y.~Guo, R.~Atadero, and J.~W. van~de Lindt, ``Streamlined bridge inspection system utilizing unmanned aerial vehicles (uavs) and machine learning,'' \emph{Measurement}, vol. 164, p. 108048, 2020.

\bibitem{UAv_earthquakBridge}
X.-W. Ye, S.-Y. Ma, Z.-X. Liu, Y.~Ding, Z.-X. Li, and T.~Jin, ``Post-earthquake damage recognition and condition assessment of bridges using uav integrated with deep learning approach,'' \emph{Structural Control and Health Monitoring}, vol.~29, no.~12, p. e3128, 2022.

\bibitem{UAVfeature}
X.~Peng, X.~Zhong, C.~Zhao, A.~Chen, and T.~Zhang, ``A uav-based machine vision method for bridge crack recognition and width quantification through hybrid feature learning,'' \emph{Construction and Building Materials}, vol. 299, p. 123896, 2021.

\bibitem{cnn}
\BIBentryALTinterwordspacing
A.~Mishra, G.~Gangisetti, Y.~E. Azam, and D.~Khazanchi, ``Weakly supervised crack segmentation using crack attention networks on concrete structures,'' \emph{Structural Health Monitoring}, vol.~0, no.~0, p. 14759217241228150, 0. [Online]. Available: \url{https://doi.org/10.1177/14759217241228150}
\BIBentrySTDinterwordspacing

\end{thebibliography}


\end{document}